\pdfoutput=1

\documentclass[11pt]{article}
\usepackage{authblk}

\usepackage{ACL2023}

\usepackage{times}
\usepackage{latexsym}
\usepackage{booktabs}
\usepackage{lscape}

\usepackage[T1]{fontenc}

\usepackage[utf8]{inputenc}

\usepackage{microtype}

\usepackage{inconsolata}

\usepackage[export]{adjustbox}
\usepackage{adjustbox}
\usepackage{multirow}
\usepackage{array}
\usepackage{hyperref}
\usepackage{subcaption}
\usepackage{float}
\usepackage{tikz}

\def\checkmark{\tikz\fill[scale=0.4](0,.35) -- (.25,0) -- (1,.7) -- (.25,.15) -- cycle;} 
\newcommand{\specialcell}[2][c]{%
  \begin{tabular}[#1]{@{}c@{}}#2\end{tabular}}

\title{IXA/Cogcomp at SemEval-2023 Task 2: Context-enriched Multilingual Named Entity Recognition using Knowledge Bases}
\author[1]{Iker García-Ferrero}
\author[1]{Jon Ander Campos}
\author[1]{Oscar Sainz}
\author[1]{\\Ander Salaberria}
\author[2]{Dan Roth}

\affil[1]{HiTZ Center - Ixa, University of the Basque Country UPV/EHU}
\affil[ ]{\{iker.garciaf,jonander.campos,oscar.sainz,ander.salaberria\}@ehu.eus}
\affil[2]{University of Pennsylvania}
\affil[ ]{danroth@seas.upenn.edu}

\begin{document}

\maketitle

\begin{abstract}
Named Entity Recognition (NER) is a core natural language processing task in which pre-trained language models have shown remarkable performance. However, standard benchmarks like CoNLL 2003 \cite{conll03} do not address many of the challenges that deployed NER systems face, such as having to classify emerging or complex entities in a fine-grained way. In this paper we present a novel NER cascade approach comprising three steps: first, identifying candidate entities in the input sentence; second, linking the each candidate to an existing knowledge base; third, predicting the fine-grained category for each entity candidate. We empirically demonstrate the significance of external knowledge bases in accurately classifying fine-grained and emerging entities. Our system exhibits robust performance in the MultiCoNER2 \cite{multiconer2-report} shared task, even in the low-resource language setting where we leverage knowledge bases of high-resource languages. 
\end{abstract}

\begin{figure}[ht!]
\centering
\includegraphics[width=0.43\textwidth]{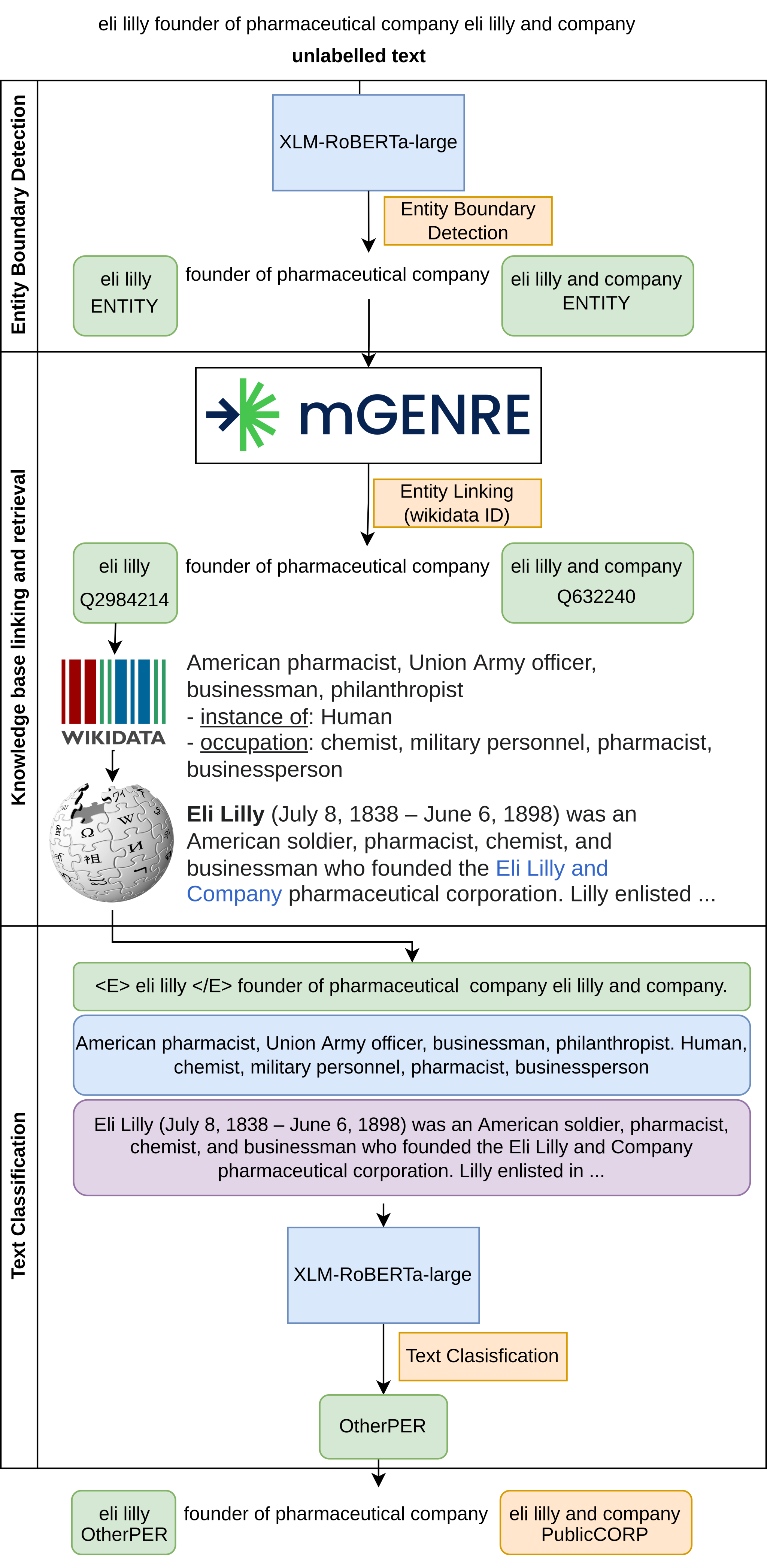}
\caption{Illustration of our fine-grained named entity recognition system.}
\label{fig:Overview}
\end{figure}

\section{Introduction}
\label{intro}

The research on the Named Entity Recognition field has mainly focused on recognizing entities from news corpora. Datasets like CoNLL 2003~\cite{conll03} or OntoNotes~\cite{hovy-etal-2006-ontonotes} have been the standard benchmarks for a long time. It is not difficult to find systems that perform over 92\% of F1 score on those datasets. However, it is known that actual models perform pretty well on relatively easy entity types (such as person names) but have been struggling to extract more complex entities~\cite{luken-etal-2018-qed, hanselowski-etal-2018-ukp}. Moreover, the evaluation setup usually involves a large overlap on the entities from the train and test sets, overestimating the systems performance. Removing such overlap resulted on a large performance drop, consequently showing that models tend to memorize rather than generalize to complex or unseen entities~\cite{meng-etal-2021-gemnet, fetahu-etal-2022-dynamic}.


Motivated by these insights, recent works have developed new and more complex datasets like MultiCoNER~\cite{malmasi-etal-2022-multiconer} or UFET~\cite{choi-etal-2018-ultra} that involve complex entity mentions with higher granularity on the type definition. To address these new challenges large language models (LLMs) are commonly used, with the hope that after being trained on billions of tokens the models become aware of the relevant entities. Despite the capacity of the models to memorize information about each individual entity, they face a harsh reality: the knowledge that a model can memorize is limited by the date on which the model was pre-trained. So, new emerging entities can be difficult to categorize for models trained years ago. Moreover, there are cases when it is not possible to identify the fine-grained category of an entity just from the context, and thus requires prior knowledge about the entity. For example: \textit{George Bernard Shaw, Douglas Fairbanks, Mary Pickford and Margaret Thatcher are some of the famous guests who stayed at the hotel.}. From the previous sentence, it is impossible to categorize \textit{George Bernard Shaw} and \textit{Margaret Thatcher} as politicians neither \textit{Douglas Fairbanks} and \textit{Mary Pickford} as artist. We hypothesize that a possible solution for this problem is to allow the models access updated Knowledge Bases, and use the updated information to infer the correct type of the entity of interest.


To overcome the mentioned challenges, in this work we present a NER approach (see Figure~\ref{fig:Overview}) that (1) identifies possible entity candidates by analyzing the input sentence structure, (2) links the candidate to an existing updated knowledge base if possible, and (3) performs the fine-grained classification using the input sentence plus the retrieved information from the KB about the entity. The approach allowed us to perform fine-grained NER with updated world knowledge when possible, and standard NER otherwise. Our code is publicly available to
facilitate the reproducibility of the results and its use in future research\footnote{\url{https://github.com/ikergarcia1996/Context-enriched-NER}}.




\section{Related Work}


Named Entity Recognition (NER) \cite{10.3115/992628.992709} is a core natural language processing task where the goal is to identify entities belonging to predefined semantic types such as organizations, products and locations. Since its inception, different approaches have been developed, from statistical machine learning methods \cite{zhou2002named, konkol2013crf,DBLP:conf/lrec/AgerriBR14} to the ones based on neural networks \cite{strubell-etal-2017-fast, xia-etal-2019-multi} and the mix of both \cite{huang2015bidirectional, chen2017improving}. Recently, the use of richer contextual embeddings computed via Transformer models \cite{vaswani2017attention, devlin2018bert, he2020deberta} have considerably improved the state-of-the-art of the task. Nevertheless, these models still have problems detecting and labelling unseen or complex entities \cite{augenstein2017generalisation, meng-etal-2021-gemnet}.

Many datasets \cite{malmasi-etal-2022-multiconer,multiconer2-data} and methods \cite{choi-etal-2018-ultra,zhou-etal-2018-zero,mengge-etal-2020-coarse} have been developed as a result of the difficulty to label complex entities. The MultiCoNER \cite{malmasi-etal-2022-multiconer} shared task focuses on detecting semantically ambiguous and complex entities in low-context settings. Almost every participant of the task used Transformer based models as their backbone model, XLM-RoBERTa \cite{conneau-etal-2020-unsupervised} being the most popular. The best two models presented for the task \cite{wang-etal-2022-damo, chen-etal-2022-ustc} show that the use of external KBs, such as Wikipedia and Wikidata, improve their results significantly.  Meanwhile, other best performing models \cite{gan-etal-2022-qtrade, pu-etal-2022-cmb, pais-2022-racai} take advantage of different data augmentation methods without relying on external KBs. The gap between the performance of the first two methods was less noticeable than the gap between the second best and the rest of approaches, implying that exploiting external KBs during training and inference is a promising research line. This difference was greater when more complex entities have to be identified. However, the overall results showed that recognizing complex or infrequent entities is still difficult \cite{multiconer-report}.

In this work we focus on the MultiCoNER2 \cite{multiconer2-data} dataset, a second version of MultiCoNER \cite{multiconer-report} with higher granularity than its predecessor. Comparing to the state-of-the-art of the first MultiCoNER task, our approach for this task also uses XLM-RoBERTa for entity boundary detection and entity classification, but differs on how we retrieve information from the external KBs. Instead of retrieving relevant text passages to classify a given entity from a query sentence \cite{wang-etal-2022-damo}, or building a Gazetteer network from Wikidata entities and their derived labels to further enrich token representations \cite{chen-etal-2022-ustc}, we directly predict Wikidata IDs from entity candidates by integrating mGENRE \cite{de2022multilingual} as our entity linking module. Previous models have already tried to link entities with external KBs \cite{tsai-etal-2016-cross,zhou-etal-2018-zero,wang2021improving}. However, mGENRE allows us to easily access both Wikidata entity descriptors and Wikipedia articles to increase our candidate's context for entity classification.

\section{System description}
Our method implements three steps. First we identify entity candidates in the input sentence. Then, we link the candidate to an existing knowledge base. Finally, we predict the fine-grained category for each entity candidate. 

\subsection{Entity Boundary Detection}
\label{sec:EntityBoundary}
\begin{figure}[t]
\centering
\includegraphics[width=0.48\textwidth]{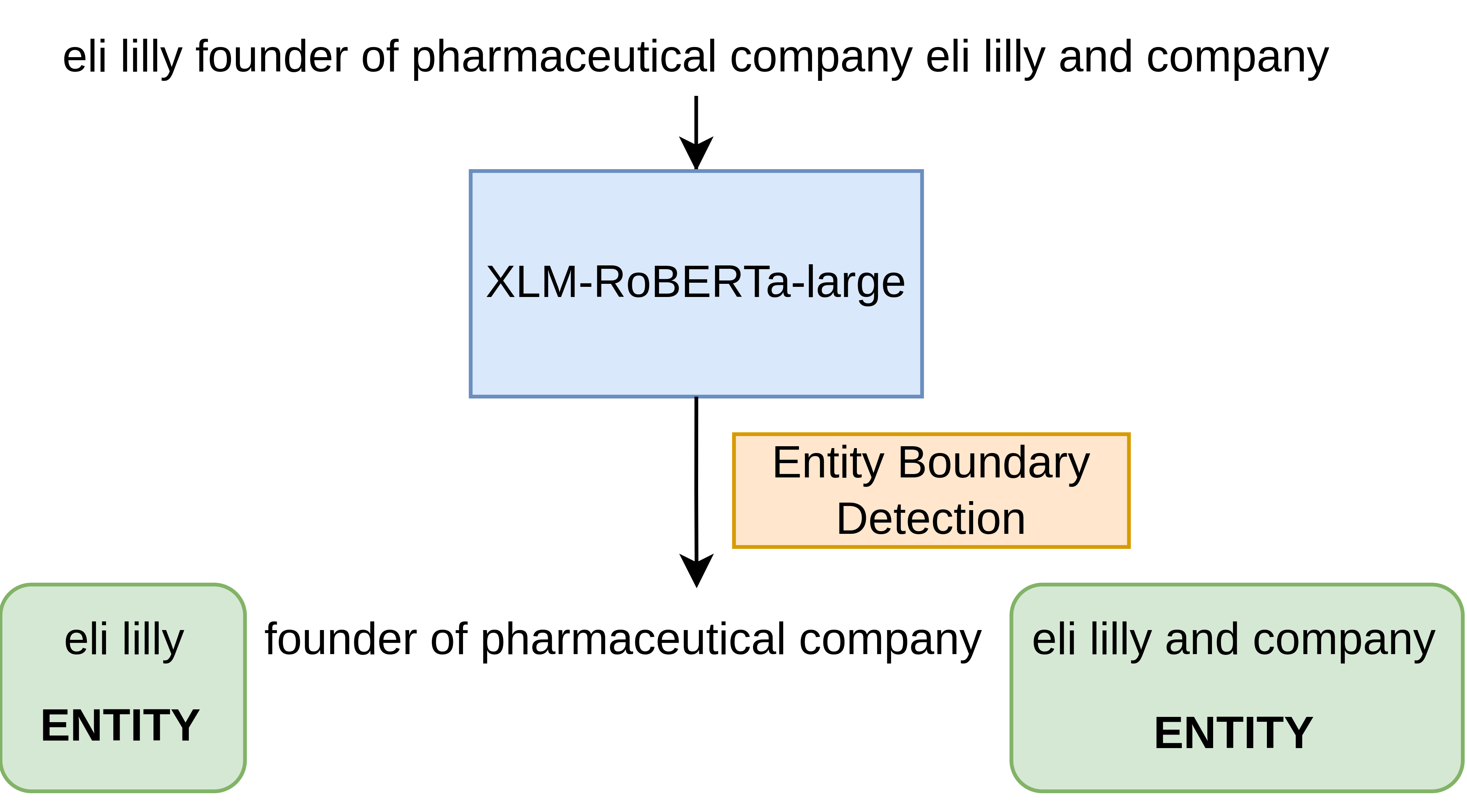}
\caption{Entity Boundary Detection. We use a XLM-RoBERTa model to predict named entity boundaries.}
\label{fig:Entity Boundary}
\end{figure}
Given unlabelled text as input, we predict named entity boundaries by analyzing the input sentence structure (Figure \ref{fig:Entity Boundary}). We treat this task as a sequence labelling task in which the model predicts if a given token is part of an entity or not ("\textit{B-ENTITY}", "\textit{I-ENTITY}","\textit{O}"). We use the multilingual XLM-RoBERTa-large model \cite{xlmr} with a token classification layer (a linear layer) on top of each token representation. Our implementation is based on the sequence labelling implementation of the Huggingface open-source library \cite{DBLP:journals/corr/abs-1910-03771}. We evaluate the model in the development set at the end of each epoch and then select the best performing checkpoint. We train five independent models and then use majority vote as the ensembling strategy at inference time.

\subsection{Entity Linking and Information Retrieval}
\label{sec:EntityRetrieval}
\begin{figure}[t]
\centering
\includegraphics[width=0.48\textwidth]{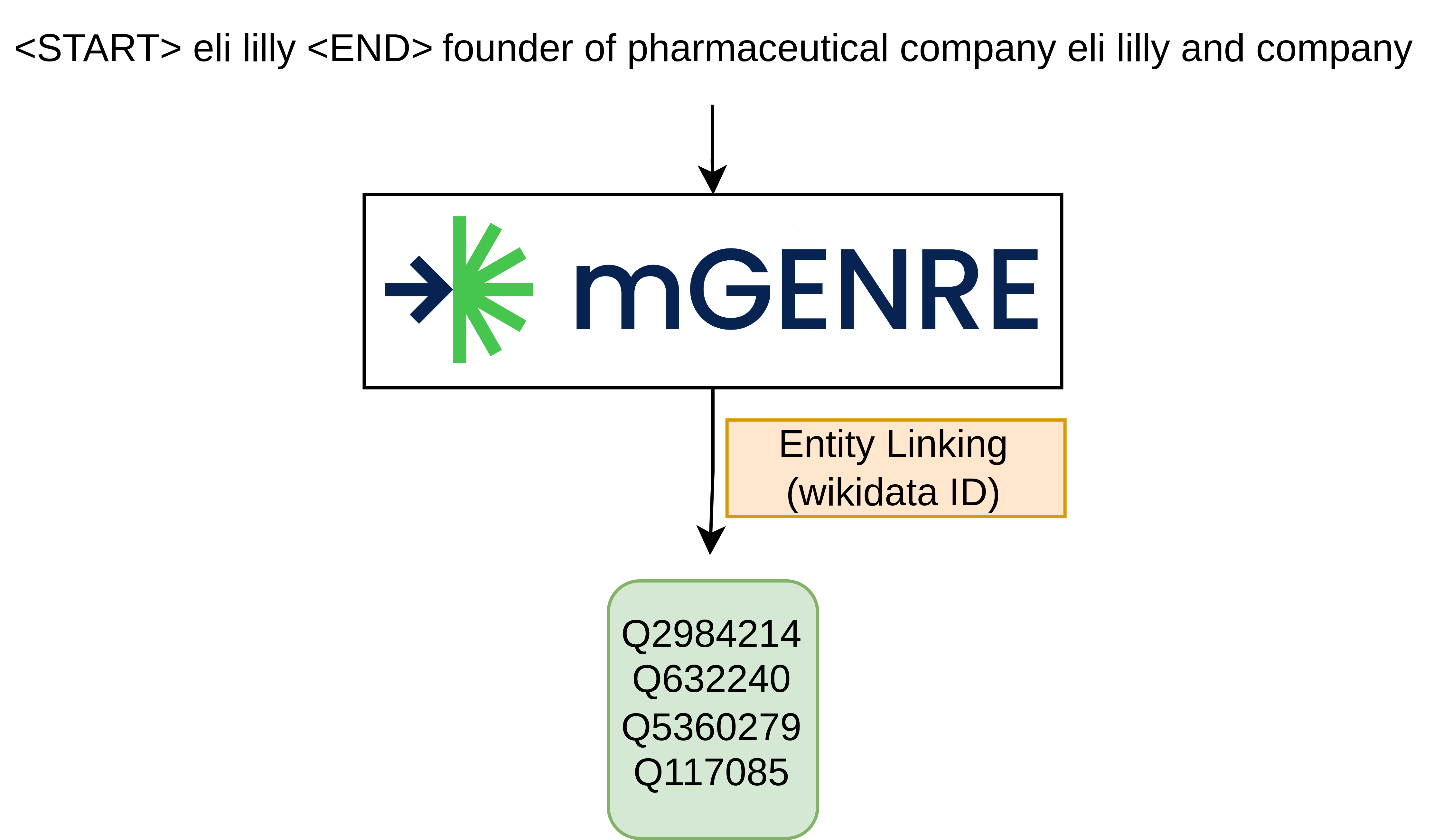}
\caption{Entity Linking step. Given a sentence and the boundaries of a named entity, we use mGENRE to predict the Wikidata ID that corresponds to the entity}
\label{fig:EntityLinking}
\end{figure}

Given a sentence and the boundaries of a named entity between the sentence, we aim to link the entity mention to its corresponding Wikidata/Wikipedia page (Figure \ref{fig:EntityLinking}). To accomplish this task, we employ the mGENRE entity linking system \cite{decao2021autoregressive,de-cao-etal-2022-multilingual}. mGENRE is a sequence-to-sequence system for Multilingual Entity Linking, which can generate entity names in over 100 languages from left to right, token-by-token in an autoregressive manner, conditioned by the context. In order to ensure that only valid entity identifiers are generated mGENRE employs a prefix tree (trie) to enable constrained beam search. mGENRE predicts both, Wikipedia page title that corresponds to the entity and the language in which the entity mention is written, and subsequently both identifiers to the Wikidata ID. In our work, we utilize the pre-trained mGENRE model provided by its authors. The authors fine-tuned an mBART \cite{DBLP:journals/corr/abs-2001-08210,lewis-etal-2020-bart} model that had been pre-trained on 125 languages using Wikipedia hyperlinks in 105 languages. For each entity, we use constrained beam search to predict the five most probable Wikidata IDs for each predicted entity from the Entity Boundary Detection step. 

\begin{figure}[t]
\centering
\includegraphics[width=0.48\textwidth]{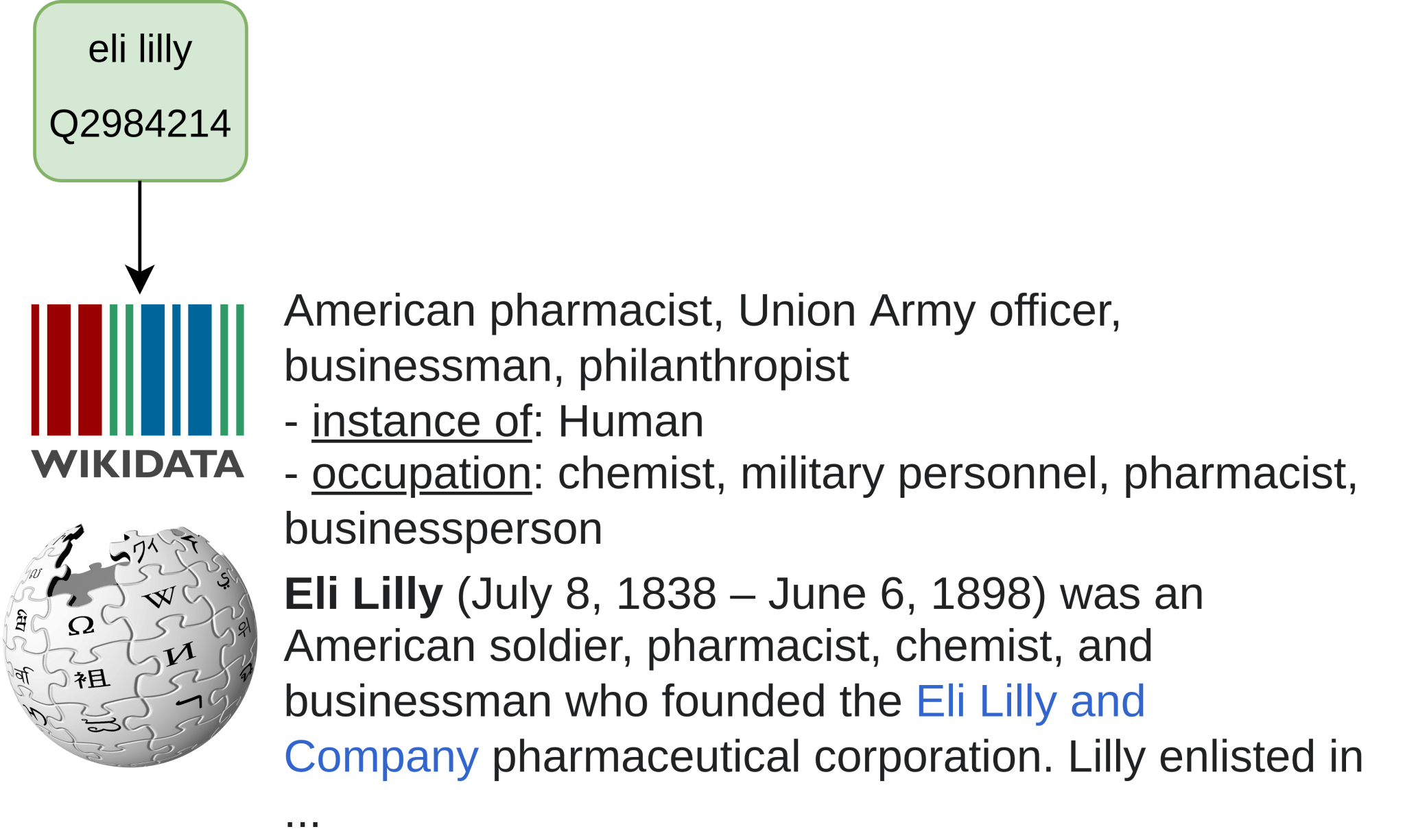}
\caption{Information Retrieval step. Given a wikidata ID, we retrieve the relevant information from the Wikidata and Wikipedia that can help us determining the correct category of the entity.}
\label{fig:Retrieval}
\end{figure}

As illustrated in Figure \ref{fig:Retrieval}, we leverage the predicted Wikidata ID to obtain the Wikidata description for the entity, as well as the contents of the \textit{instance\_off} and \textit{occupation} arguments. In the multilingual track, we also retrieve the \textit{subclass\_off} argument. Due to time constraints, we were unable to retrieve this argument in the other tracks. As Wikidata pages contain links to their corresponding Wikipedia webpages, we also retrieve the Wikipedia summary. These summaries are generally more detailed than the Wikidata descriptions.

We initiate the retrieval process by querying the Wikidata ID with the highest probability, as predicted by mGENRE. If the predicted ID corresponds to a page that has been deleted, is empty, or is a list/disambiguation page, we discard the ID and proceed to the next most probable predicted ID. We found that, while Wikidata descriptions and arguments are mostly available in English for the entities in MultiCoNER, this is not the case for other languages such as Bangla or Farsi. Similarly, a large number of entities have a Wikipedia page available in English but not in other languages. For this reason, we always retrieve the Wikidata description and arguments, as well as the Wikipedia summary, in English.

\subsection{Entity Category Classifier}
\label{sec:EntityClass}
\begin{figure}[t]
\centering
\includegraphics[width=0.48\textwidth]{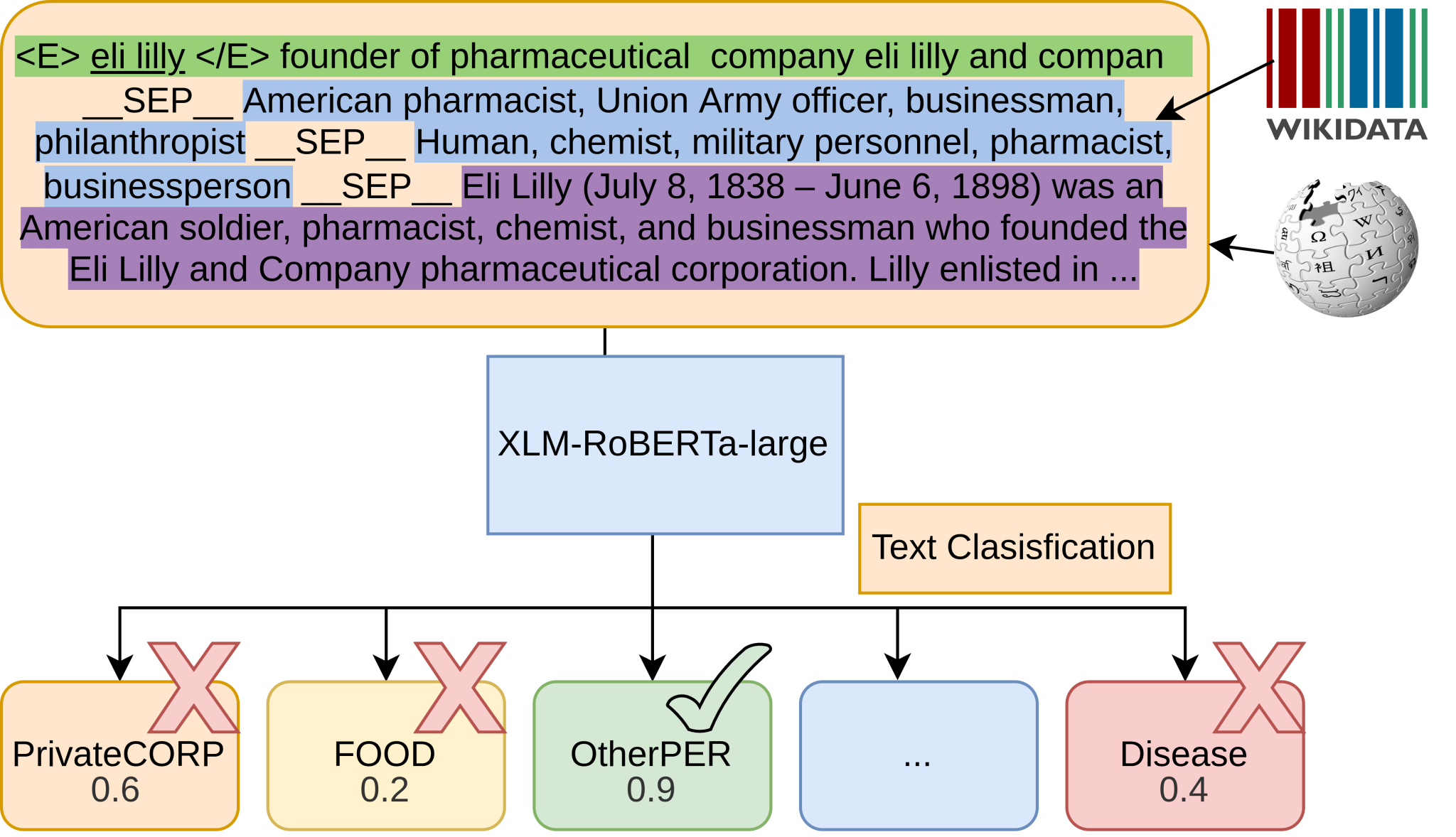}
\caption{Given the boundaries of a named entity and the information retrieved from Wikipedia/Wikidata, we use a XLM-RoBERTa model to predict the correct category for the entity.}
\label{fig:TextClass}
\end{figure}

In our final step, we aim to classify the entity candidates into fine-grained categories by combining all the knowledge we have. We create a text input by concatenating the sentence where the entity is annotated with an HTML-style markup, the wikidata description, a list of all  retrieved arguments, and the Wikipedia summary. We use the special token \textit{\_\_SEP\_\_} to delimit each piece of information. If we fail to retrieve a Wikidata or Wikipedia page, we use the text "No Wikidata/Wikipedia summary found," and the model predicts the category solely based on the context of the entity.

We use the multilingual XLM-RoBERTa-large model \cite{xlmr} with a text classification layer (a linear layer) on top of the first token. We implement this using the Huggingface open-source library's text classification implementation \cite{DBLP:journals/corr/abs-1910-03771}. During training, we generate the train and development datasets using the gold entity boundaries. During inference, we predict the categories for the entity boundaries computed in the entity boundary detection step. We evaluate the model at the end of each epoch and choose the model with the highest performance on the development set. We train five independent models and merge the predictions at inference using majority voting.

\section{Experimental Setup and Dataset}
We use the official MultiCoNER2 dataset \cite{multiconer2-data} in all tracks to train our entity boundary and entity classification models. To query external knowledge, we utilize the MediaWiki API to retrieve data from the live version of Wikipedia/Wikidata.
MultiCoNER2 consists of two substacks, the monolingual task with 12 monolingual datasets and a multilingual task that contains data from all languages. For 7 languages (EN, ZH, IT, ES, FR, PT, SV) some sentences were corrupted with noise, either on context tokens or entity tokens. The results of the shared task are evaluated based on entity-level macro F1 scores, where all label categories contribute equally to the final score, regardless of the number of entities labeled with each category.
The dataset consists of 36 pre-defined fine-grained categories which are grouped into 6 coarse-grained categories: Medical entities, Locations, Creative Works, Groups, Persons and Products. 
A description of the hyper-parameters used, dataset stats and hardware used in available in the Appendix.

\section{Results}
\label{sec:results}

\begin{table*}[htb]
    \centering
 \adjustbox{max width=\linewidth}{

\begin{tabular}{c|ccccccccccccc|c}
\toprule
Team Name & EN & ES & SV & UK & PT & FR & FA & DE & ZH & HI & BN & IT & MULTI & AVG \\ 
\midrule
XLM-RoBERTa-large & 59.85 & 62.60 & 65.76 & 65.67 & 61.09 & 62.98 & 57.17 & 62.86 & 51.89 & 68.06 & 65.97 & 65.17 & 69.12 &  62.93\\ 
\midrule
NetEase & - & - & - & - & - & - & - & - & \textbf{84.05} & - & - & - & - & - \\
USTC-NELSLIP & 72.15 & 74.44 & 75.47 & 74.37 & 71.26 & 74.25 & 68.85 & 78.71 & 66.57 & \textbf{82.14} & 80.59 & 75.70 & 75.62 & 74.72\\
PAI & 80.00 & 71.67 & 72.38 & 71.28 & 81.61 & 86.17 & 68.46 & \textbf{88.09} & 74.87 & 80.96 & \textbf{84.39} & 84.88 & 77.00 & 78.59\\
DAMO-NLP & \textbf{85.53} & \textbf{89.78} & \textbf{89.57} & \textbf{89.02} & \textbf{85.97} & \textbf{89.59} & \textbf{87.93} & 84.97 & 75.98 & 78.56 & 81.60 & \textbf{89.79} & \textbf{84.48} & \textbf{85.59} \\ 
\midrule
Our System & 72.82 & 73.81 & 76.54 & 75.25 & 72.28 & 74.52 & 69.49 & 80.35 & 64.86 & 79.56 & 78.95 & 74.67 & 78.17 & 74.71 \\ 
\bottomrule
\end{tabular}
}
    \caption{Our system macro-F1 score for all the tracks compared with our baseline and the systems that achieved the best results.}
    \label{tab:SharedTaskResults}
\end{table*}

Our team participated in all the 12 monolingual tracks and the multilingual track. More than 35 team participated in the shared task. We achieved the \textbf{3rd} positions in four tracks. The shared task results computed by the organizers are shown in Table \ref{tab:SharedTaskResults}. We first compare our system with an XLM-RoBERTa-large \cite{xlmr} baseline. For this baseline, we added a token classification layer (linear layer) on top of each token representation of an XLM-RoBERTa-large model and fine-tuned the model to directly predict the fine-category for each token. The results were much lower than the ones achieved with our system, which confirms our hypothesis that external knowledge is needed to predict the entity categories.


We also compare our system with the performance of systems from others teams that got the first position in any of the tracks. 
Our system was most competitive in Hindi and Bangla, as well as the multilingual track. Hindi and Bangla are by far the languages with the smallest Wikipedia's, ranked 59th and 63rd by the number of articles respectively.\footnote{\url{https://meta.wikimedia.org/wiki/List_of_Wikipedias}} We attribute this to our decision to retrieve the external knowledge from the English Wikipedia/Wikidata for all the tracks. Because we predict a wikidata ID, we are able to retrieve knowledge in any language in which it is available. This enables our system to be used for low-resource languages for which no large, up-to-date knowledge base is available, thanks to being able to link the entities found to a knowledge base in high-resource languages. On the other hand, our system is not as competitive in very high-resource languages, such as English and Chinese.

\subsection{Clean vs Noisy Performance}

\begin{table}[t] 
    \centering
 \adjustbox{max width=\linewidth}{

\begin{tabular}{c|cc}
\toprule
Track & Clean & Noisy \\
\midrule
EN & 76.64 & 64.36 \\
ES & 77.65 & 66.09 \\
SV & 80.75 & 68.69 \\
PT & 76.00 & 65.54 \\
FR & 78.60 & 65.81 \\
ZH & 70.35 & 48.37 \\
IT & 78.16 & 67.66 \\ 
\bottomrule
\end{tabular}
}

    \caption{Our system macro-F1 score in the clean and noisy data from the test sets as computed by the organizers}
    \label{tab:CleanVSNoisy}
\end{table}

In seven languages (EN, ZH, IT, ES, FR, PT, SV), some sentences contained noise, either on context tokens or entity tokens. Table \ref{tab:CleanVSNoisy} shows the results of our system on the set of sentences that were not corrupted and the set of corrupted sentences. We observed a significant drop in performance when noise was introduced in the sentences, especially for the Chinese track. We believe that this is because our entity linking step depends too much on the entity itself and does not sufficiently take into account the context in which the entity appears. Improving our system's ability to consider contextual information is something we need to work on.

\subsection{Ablation Study}
In this section, we evaluate the different steps in our system and check how different design decisions affect performance.

\paragraph{How important is external knowledge to classify entities?}

%
%
%

We aim to measure the importance of retrieving external knowledge for accurate entity classification in the MultiCoNER dataset. To achieve this, we train our text classification system on the training split and evaluate its performance on the development data. We use the gold labeled entity spans for both splits. We assess the relevance of external knowledge by varying the amount of external knowledge used in the evaluation.
Table \ref{tab:ContextVsPerformance} shows the results obtained using different combinations of external knowledge, including using only the entity in its context (no external knowledge). As we observed in Table \ref{tab:SharedTaskResults}, the results without external knowledge are very poor. The most relevant external knowledge is the content of the Wikidata arguments. However, as we provide more external knowledge to the model, the performance improves. These results also demonstrate that our system achieves remarkable results in classifying entities into fine-grained categories when the correct entity spans are provided, indicating that the entity boundary detection step hinders the performance of our system. The entity classification confusion matrix for the English development set is available in Appendix \ref{sec:ConfusionMatrix}.

 \begin{table}[t]
    \centering
 \adjustbox{max width=\linewidth}{

\begin{tabular}{cccc|ccc}
\toprule
\specialcell{Initial \\ Context} & \specialcell{Wikidata \\ Description} & \specialcell{Wikidata \\ Arguments} & \specialcell{Wikipedia \\ Summary} & EN & ZH \\ 
\midrule
\checkmark & & & & 79 & 81 \\ 
\checkmark & \checkmark & & & 82 & 84 \\ 
\checkmark & & \checkmark & & 88 & 88 \\ 
\checkmark & & & \checkmark & 86 & 86 \\ 
\checkmark & \checkmark & \checkmark & & 89 & 86 \\ 
\checkmark & \checkmark & \checkmark & \checkmark & 90 & 89 \\ 
\bottomrule
\end{tabular}
 
}

    \caption{Our system macro-F1 score in the development split when using gold entity boundaries and different amount of external knowledge}
    \label{tab:ContextVsPerformance}
\end{table}

\paragraph{Entity boundary detection performance}
\begin{table}[t] 
    \centering
     \adjustbox{max width=.8\linewidth}{
    
    \begin{tabular}{c|cccc}
    \toprule
    & EN & ZH & Multi \\
    \midrule
    Entity Boundary Detection & 90.05 & 90.49 & 90.52 \\ 
    \bottomrule
    \end{tabular}
}

    \caption{F1-score of the Entity Boundary Classification Step in the development split}
    \label{tab:EntityBoundaryPerformance}
\end{table}

Our initial hypothesis was that external knowledge is required to accurately classify entities into fine-grained categories. However, we also hypothesized that detecting the boundaries of named entities does not require external knowledge. For instance, it is possible to identify that a sequence of words corresponds to the name of a person without needing to know anything about that person. Table \ref{tab:EntityBoundaryPerformance} shows the performance the entity boundary detection step in the development dataset, which does not utilize any external knowledge. We were able to identify a significant proportion of the entities in the dataset. However, the results were lower than those of the entity classification step in Table \ref{tab:ContextVsPerformance}. This indicates that the entity boundary detection step is the weakest point in our system. We believe that incorporating external knowledge into this step would enhance the results.

\section{Conclusion}

We have developed a system that identifies potential entity candidates and leverages external knowledge from an up-to-date knowledge base to classify them into a set of predefined fine-grained categories, addressing the challenges posed by temporal knowledge and unknown entities. Our results demonstrate exceptional performance in classifying entities into fine-grained categories, underscoring the need for external knowledge in accurately classifying entities in the MultiCoNER dataset. However, our current entity boundary detection step does not incorporate external knowledge, which we plan to improve in the future.

Furthermore, our system exhibits promising results for low-resource languages where a comprehensive Wikipedia is unavailable. By linking the identified entities to a knowledge base in a high-resource language, it can be used to process low-resource languages. In future work, we intend to integrate all the steps and insights gained from our current system into a single end-to-end model. Although each step performs well when evaluated independently, the current pipeline multiplies errors at each step, which we aim to address in the integrated model.




\section*{Acknowledgements}
This work has been partially supported by the HiTZ center and the Basque
Government (Research group funding IT-1805-22). We also acknowledge the funding from the following projects: 
(i) DeepKnowledge (PID2021-127777OB-C21) MCIN/AEI/10.13039/501100011033 and ERDF A way of making Europe;
(ii) DeepR3 (TED2021-130295B-C31) by MCIN/AEI/10.13039/501100011033 and
EU NextGeneration programme EU/PRTR. 
(iii) IARPA BETTER Program contract No. 2019-19051600006 (ODNI, IARPA).

Iker García-Ferrero, Oscar Sainz and Ander Salaberria are supported by doctoral grants from the Basque Government (PRE\_2022\_2\_0208, RE\_2022\_2\_0110 and PRE\_2022\_2\_0219, respectively). Jon Ander Campos enjoys a doctoral grant from the Spanish MECD (FPU18/01271).

\bibliography{custom}
\bibliographystyle{acl_natbib}

\appendix
\label{sec:appendix}

 \section{Dataset details}

\begin{figure*}[htb]
\centering
\includegraphics[width=0.70\textwidth]{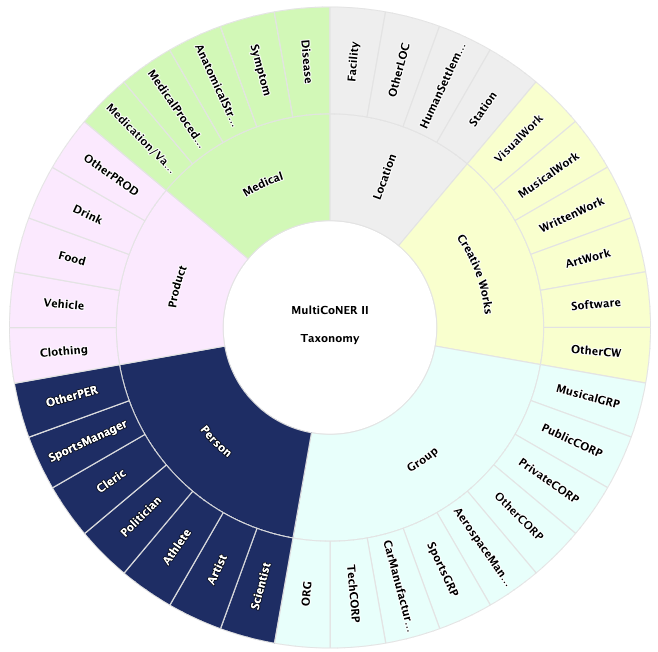}
\caption{MultiCoNER2 Taxonomy}
\label{fig:Taxonomy}
\end{figure*}

 \begin{figure*}[htb]
\centering
\includegraphics[width=0.95\textwidth]{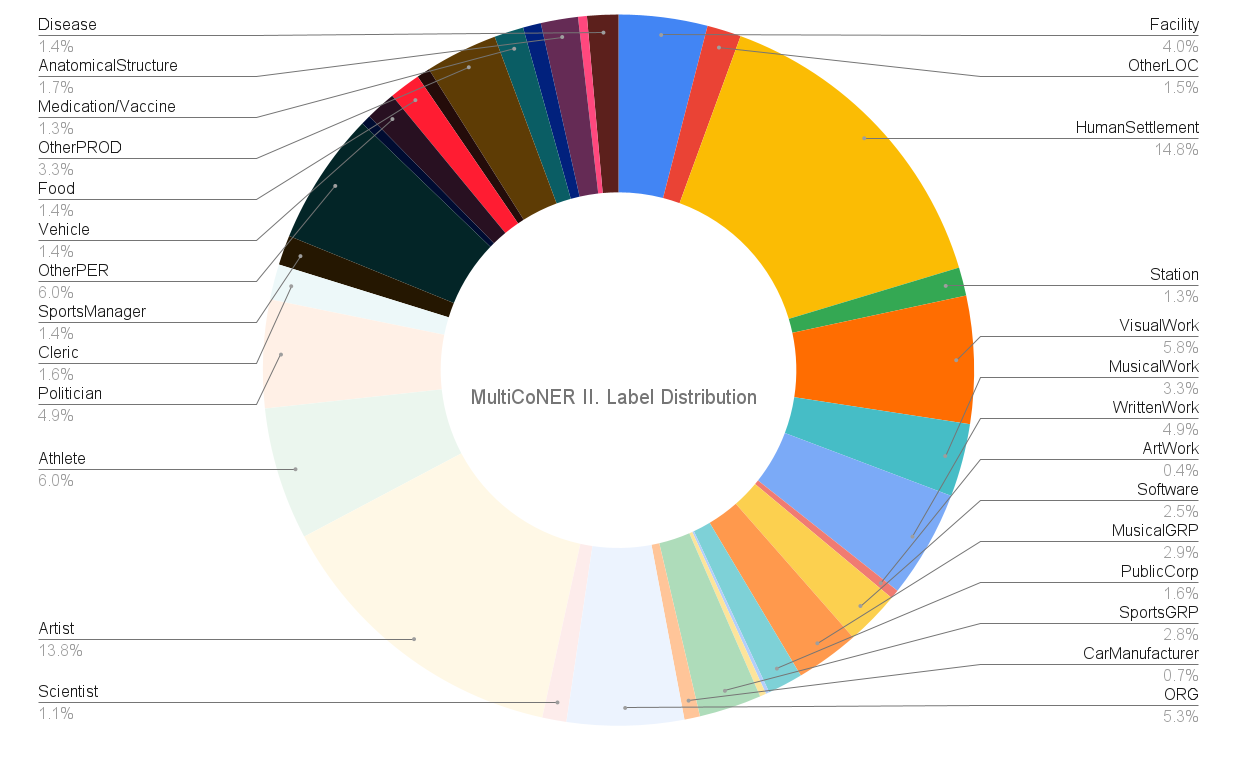}
\caption{MultiCoNER2 label distribution computed as the sum of the label count for every language and split}
\label{fig:LabelDistribution}
\end{figure*}

 We use the MultiCoNER2 dataset in all our experiments \cite{multiconer2-data}. The number of sentences for each language and dataset split is available in Table \ref{tab:DatasetSize}. The dataset Taxonomy is visualized in Figure \ref{fig:Taxonomy}. The label distribution is visualized in Figure \ref{fig:LabelDistribution}.

\section{Hyper parameters}

\subsection{Entity Boundary Detection}
We use the XLM-RoBERTa-large model in all our experiments. We train the model for 8 epochs, with a batch size of 16, AdamW optimizer \cite{DBLP:conf/iclr/LoshchilovH19} with 2e-5 learning rate. We use a max sequence size of 256 tokens, larger sequences are truncated to the first 256 tokens. The rest of hyper parameters are the default values for the sequence labelling implementation of the Huggingface library \footnote{\url{https://github.com/huggingface/transformers/tree/main/examples/pytorch/token-classification}}. 
As described in \ref{sec:EntityBoundary} we evaluate the model in the development set at the end of each epoch and then select the best performing checkpoint. We train five independent models and then use majority vote as the ensembling strategy at inference time.  

\subsection{Entity Linking and Information Retrieval}
We use the mGENRE implementation \footnote{\url{https://github.com/facebookresearch/GENRE/tree/main/examples_mgenre}} in the fairseq library \cite{DBLP:conf/naacl/OttEBFGNGA19}. We use the prefix tree provided buy the authors and marginalization. This constrains the model to only generate valid identifiers. 
We retrieve data from Wikipedia using the pymediawiki \footnote{\url{https://github.com/barrust/mediawiki}} wrapper and parser for the MediaWiki API. We use Wikidata client library for Python \footnote{\url{https://github.com/dahlia/wikidata}} to retrieve knowledge from Wikidata. 

\subsection{Entity Category Classifier}
We use the XLM-RoBERTa-large model in all our experiments. We train the model for 8 epochs, with a batch size of 16, AdamW optimizer \cite{DBLP:conf/iclr/LoshchilovH19} with 2e-5 learning rate. We use a max sequence size of 256 tokens. The rest of hyper parameters are the default values for the text classification implementation of the Huggingface library \footnote{\url{https://github.com/huggingface/transformers/tree/main/examples/pytorch/text-classification}}. 
As described in \ref{sec:EntityClass} we evaluate the model in the development set at the end of each epoch and then select the best performing checkpoint. We train five independent models and then use majority vote as the ensembling strategy at inference time.

\section{Hardware used}
We perform all our experiments using a single NVIDIA A100 GPU with 80GB memory. The machine used has two AMD EPYC 7513 32-Core Processors and 512GB of RAM. Although our system can also run on GPUs with 24GB of VRAM.

\section{Text Classification Confusion Matrix}
\label{sec:ConfusionMatrix}
\begin{figure*}[htb]
\centering
\includegraphics[width=0.98\textwidth]{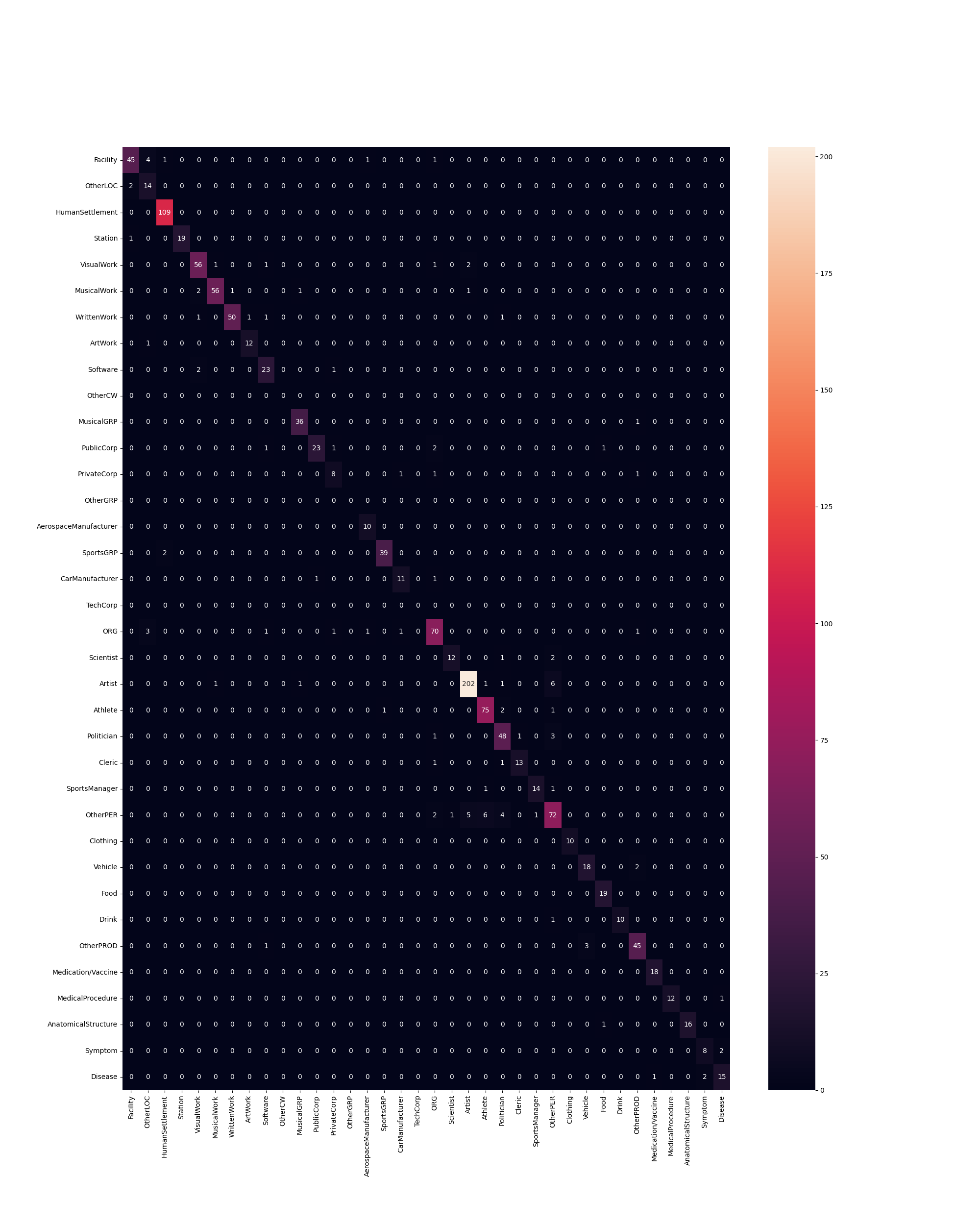}
\caption{text classification confusion matrix for the English development test using gold labelled spans}
\label{fig:ConfussionMatrix}
\end{figure*}

We show the text classification confusion matrix (Figure \ref{fig:ConfussionMatrix}) for the English development test using the gold labelled spans to better understand which type of entities are the most difficult ones to classify for our model. As we can see in the matrix, the entities related to persons are the most ambiguous and difficult to classify. 

\begin{table}[htb]
    \centering
 \adjustbox{max width=\linewidth}{

\begin{tabular}{c|cc}
\toprule
Lang & Split & Sentences \\
\midrule
\multirow{3}{*}{ BN } & train & 9,708 \\
& dev & 507 \\
& test & 19,859 \\
\midrule
\multirow{3}{*}{ DE } & train & 9,785 \\
& dev & 512 \\
& test & 20,145 \\
\midrule
\multirow{3}{*}{ EN } & train & 16,778 \\
& dev & 871 \\
& test & 249,980 \\
\midrule
\multirow{3}{*}{ ES } & train & 16,453 \\
& dev & 854 \\
& test & 246,900 \\
\midrule
\multirow{3}{*}{ FA } & train & 16,321 \\
& dev & 855 \\
& test & 219,168 \\
\midrule
\multirow{3}{*}{ FR } & train & 16,548 \\
& dev & 857 \\
& test & 249,786 \\
\midrule
\multirow{3}{*}{ HI } & train & 9,632 \\
& dev & 514 \\
& test & 18,399 \\
\midrule
\multirow{3}{*}{ IT } & train & 16,579 \\
& dev & 858 \\
& test & 247,881 \\
\midrule
\multirow{3}{*}{ PT } & train & 16,469 \\
& dev & 854 \\
& test & 229,490 \\
\midrule
\multirow{3}{*}{ SV } & train & 16,363 \\
& dev & 856 \\
& test & 231,190 \\
\midrule
\multirow{3}{*}{ UK } & train & 16,429 \\
& dev & 851 \\
& test & 238,296 \\
\midrule
\multirow{3}{*}{ ZH } & train & 9,759 \\
& dev & 506 \\
& test & 20,265 \\
\midrule
\multirow{3}{*}{ MULTI } & train & 170,824 \\
& dev & 8,895 \\
& test & 358,668 \\
\bottomrule
\end{tabular}

}
    \caption{Number of sentences for each file in the dataset}
    \label{tab:DatasetSize}
\end{table}


\end{document}